# Entropy-Based Search Algorithm for Experimental Design

N. K. Malakar[a] and K. H. Knuth[a,b]

[a]Department of Physics
[b]Department of Informatics
University at Albany (SUNY), Albany NY USA

**Abstract.** The scientific method relies on the iterated processes of inference and inquiry. The inference phase consists of selecting the most probable models based on the available data; whereas the inquiry phase consists of using what is known about the models to select the most relevant experiment. Optimizing inquiry involves searching the parameterized space of experiments to select the experiment that promises, on average, to be maximally informative. In the case where it is important to learn about each of the model parameters, the relevance of an experiment is quantified by Shannon entropy of the distribution of experimental outcomes predicted by a probable set of models. If the set of potential experiments is described by many parameters, we must search this high-dimensional entropy space. Brute force search methods will be slow and computationally expensive. We present an entropy-based search algorithm, called nested entropy sampling, to select the most informative experiment for efficient experimental design. This algorithm is inspired by Skilling's nested sampling algorithm used in inference and borrows the concept of a rising threshold while a set of experiment samples are maintained. We demonstrate that this algorithm not only selects highly relevant experiments, but also is more efficient than brute force search. Such entropic search techniques promise to greatly benefit autonomous experimental design.

## INTRODUCTION

The scientific method of learning relies on the iterated processes of hypothesis generation, inference and inquiry. Given the learning goal, hypotheses, usually in the form of parameterized models, are generated. We use the available data to learn about our hypotheses by computing probabilities to quantify what we know about the model parameters. What is known about the models, or equivalently, what remains unknown about the models, can be used to identify and select the most relevant experiment for the next cycle of data collection and learning. Considering a fixed set of hypotheses, the iterated processes of inference followed by inquiry comprise what is known as data driven or active learning.

Our ultimate goal is focused on designing autonomous intelligent systems that collect their own data while they learn. Such a system will rely heavily on automating the processes of inference and inquiry and its performance will depend upon its ability to explore adaptively and learn actively. In general, without prior information about the specific learning problem, such a system is required to take the greedy approach where the main goal is to choose each data collection step so that it is maximally informative toward the experimental goal. Given a data set and the inferences drawn from that data, the problem of future inquiry can be viewed as a search problem through the space of questions where we seek the maximally informative question given the both the available data and the learning goal. For experimental questions characterized by more than a few parameters, this promises to become a computationally expensive process. In this paper we propose an efficient sampling algorithm to search the space of potential experimental questions to identify the one which maximizes the potential information gain.

## AUTOMATED EXPERIMENTAL DESIGN

We begin with a set of hypotheses described by parameterized models. The inference phase consists of updating our beliefs about the parameterized models based upon the collected data. Given the data and a prior probability we use Bayes' theorem to make inferences about the model parameters. The result is encoded by the posterior probability:

$$p(\theta/\mathbf{D}) = p(\theta) \frac{p(\mathbf{D}|\theta)}{p(\mathbf{D})} \qquad //I \qquad (1)$$

where the data are represented by $\mathbf{D}$ and parameters are represented by $\theta$. We have adopted John Skilling's convention of writing the prior information $I$ as the generally accepted term in the conditionals.

Inquiry is the natural extension of inference. Critical to inquiry is the uncertainty in our inferences. We represent this uncertainty implicitly by sampling probable model instantiations from the posterior probability. One then can query the set of sampled models to discover what each one predicts will be measured for any proposed experiment. The resulting distribution of predicted measurements represents our uncertainty about the results of any proposed experiment.

Making a decision based upon the knowledge about the inferred models is described under the decision-theoretic approach to Bayesian experimental design. Central to this approach is the utility function, which represents the joint effect of constraints relevant to the decision making process. A decision should be made by maximizing some expected utility [7, 8, 1]. In the decision-theoretic literature, such a utility is also known as merit of consequences. Given a choice of a particular utility function, it is then straightforward to automate the decision process.

Given the previously recorded data $\mathbf{D}$, we can make inferences about the model $M$ and associated parameters $\theta$ thereby allowing us to get the predictive distribution, or likelihoods $p(d/\mathbf{D}, e, M)$ of data $d$. We would like to decide on the best experiment out of all candidate experiments "$e$" which promises data "$d$". The expected utility for conducting such an experiment is:

$$EU_e = \sum_d p(d/\mathbf{D}, e, M) U_{d,e}, \qquad (2)$$

where $U_{d,e}$ denotes the utility of conducting an experiment $e$ which promises data $d$. By using the product rule, the expected utility of an experiment can be written as:

$$EU_e = \sum_\theta \sum_d p(d/\mathbf{D}, \theta, e, M) p(\theta/\mathbf{D}, e, M) U_{d,e}. \qquad (3)$$

To select the best experiment "$e$", that provides the greatest expected gain in information, we use the Shannon information for the utility function

$$U_{d,e} = \sum_\theta p(\theta/d, \mathbf{D}, e, M) \log p(\theta/d, \mathbf{D}, e, M), \qquad (4)$$

One can show that the optimal experiment that maximizes the expected utility given by (3) along with (4) can be found by maximizing the entropy of the distribution of potential measurements [9, 4, 6]

$$\hat{e} = \arg\max \left( C - \sum_d p(d/\mathbf{D}, M) \log p(d/\mathbf{D}, M) \right). \qquad (5)$$

Therefore, in the case that we are interested in learning about all the relevant model parameters, the experiment resulting in a distribution of potential measurements with the greatest entropy promises to be the most informative [11, 9, 15, 5, 16, 6]. Note that if we are interested in only a subset of parameters, then the optimal experiment is found by computing the mutual information [10]. The result is that we use the posterior probability to select a sample of possible models. Each model is then queried to produce a set of predictions. Taken together the entropy of this set of potential measurements is used to select an optimal experiment. The net result is that the optimal experiment is the one with the most uncertain outcomes.

The next section introduces an algorithm that searches for an optimal experiment in the entropy space.

## NESTED ENTROPY SAMPLING

We propose an entropy-based search algorithm called Nested Entropy Sampling (NES) which is inspired by the nested sampling algorithm [13, 14]. Implementation of nested sampling involves sampling from the prior distribution and imposing ever-evolving hard likelihood constraints. However, here we are exploring an entropy space (*H-space*) and propose to evolve a hard-entropy constraint. The goal is to find the global maxima in the entropy space by utilizing the technique of exploration and nested contraction of the search domain. This will avoid the need to compute the entropy for every candidate experiment, which will greatly speed up the adaptive exploration process.

For experiments parameterized by a number of parameters, $p$, the search needs to be done in a $p$-dimensional space. In NES, for convenience, the search space is divided into a discrete number of cells so that each experiment is indexed by a set of integers. We select *N* random sample experiments, $\{e_1, e_2, ... e_N\}$, and compute the entropy for these selected experiments $\{H(e_1), H(e_2), ... H(e_N)\}$. This provides the starting point for the exploration stage.

**TABLE 1: NESTED ENTROPY SAMPLING PSEUDO CODE**

| |
|---|
| INPUT: posterior samples from the inference phase |
| SET UP: Generate a set of *N* sample experiments randomly and compute the entropy *H* for each. <br> WHILE samples have different entropy values <br>   Select the sample *e\** from the set that has the least entropy, denoted *H\**. <br>   Generate a trial experiment $e_{trial}$ by selecting another sample at random from the set <br>   EXPLORE LOOP <br>     Explore by varying the parameters of $e_{trial}$ and computing the new value of $H_{trial}$. <br>     Accept the trial if *H* > *H\** otherwise reject it. <br>     Monitor acceptance range and change exploration step size <br>   Replace *e\** with $e_{trial}$ <br> END WHILE |
| OUTPUT: prescription of the optimal experiment. |

Given the set of sample experiments, we select a sample *e\** with the least entropy and call it *H\**

$$H^* = min\{H(e_1), H(e_2), ... H(e_N)\}. \qquad (6)$$

We then copy a sample from the set with *H* > *H\** to generate a trial experiment $e_{trial}$ and replace the discarded experiment *e\** by $e_{trial}$ so that the number of samples is conserved. Exploration of the entropy space is performed by varying the trial sample a set number of times. A trial sample is accepted as long as its entropy $H_{trial}$ > *H\**. The value of *H\** defines an entropy threshold that implicitly defines a search boundary that can be freely explored. We record entropy values of the visited cells so that, when necessary, the entropy value can be retrieved easily from a look-up table instead of being re-computed. This saves a good number of computations. Iteration of the exploration step continues by contracting a nested set of search domain boundaries defined by successive values of *H\** until all of the samples have converged to a single maximum entropy value $H_{max}$. Note that the search space may be multimodal, which corresponds to multiple peaks in the space each representing locally optimal experiments. Therefore, the algorithm has the potential to converge at multiple peaks suggesting a set of equally optimal experiments.

Since we are interested in finding an optimal experiment, we do not compute the weighted mean of the experimental parameters. If the algorithm suggests multiple optimal experiments, we select a single experiment either at random from the optimal set or according to an additional cost function. The additional cost function may be the cost of conducting the experiment, time factor or energy budget necessary to conduct such experiments.

## Compression Efficiency

To calculate the efficiency of the *NES* algorithm, we compute the ratio of the number of times the algorithm was required to compute the entropy versus the total number of experiments. We call this the *Compression Efficiency* (*CE*)

$$CE = n/m, \qquad (7)$$

where *n* is the total number of candidate experiments and *m* is the number of times that the entropy was computed during the *NES* run time. A brute force search method requires *n* computations, which results in compression efficiency of one, whereas more efficient algorithms will result in compression efficiencies greater than one.

## EXPERIMENTS WITH MIXTURES OF GAUSSIANS

To evaluate our algorithm, we tested it on a large number of two-dimensional entropy landscapes defined by a mixture of randomly generated Gaussians

$$H(x,y) = \sum_i K_i \, Exp\left[-\frac{1}{2}\left\{A_i(x-x_i)^2 + B_i(y-y_i)^2 + 2C_i(x-x_i)(y-y_i)\right\}\right] \qquad (8)$$

where $K_i$ denotes the amplitude of the $i^{th}$ Gaussian. Note that each point in the *H-space* given by (8) corresponds to an experiment effectively parameterized by the coordinates ($x_i$, $y_i$). The goal is to locate the coordinates of the maximum peak in the Gaussian entropy landscape.

We explored a wide variety of landscapes each generated by the mixture of Gaussians model and indexed according to the number of Gaussians used to define the landscape. For a given number of Gaussians, we randomly generated entropy landscapes and calculated the mean compression efficiency for various numbers of runs. Figure 1 shows one such entropy landscape defined by mixture of seven Gaussians. The "+" symbols have been used to denote which coordinates were sampled during convergence to the global peak.

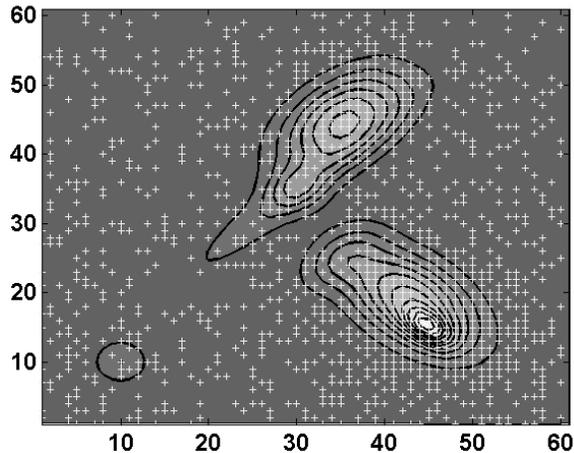

**FIGURE 1**. Convergence of the *NES* algorithm in an entropy landscape characterized by seven Gaussians. The '+' signs indicate coordinates that have been explored.

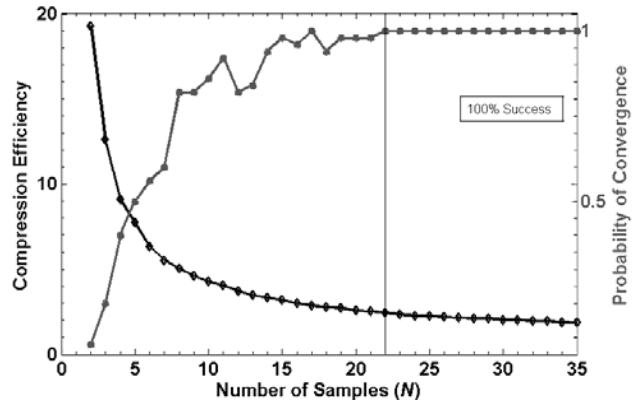

**FIGURE 2**. The mean compression efficiency decreases as a function of the number of samples $N$; whereas the probability of convergence increases with number of samples. This graph gives some idea about the minimum number of samples needed for optimally successful search in this scenario.

We tested the algorithm by varying the number of exploring samples $N$ for the Gaussian landscape and running it 100 times for each value of $N$. Figure 2 shows both the mean compression efficiency and the probability of successful convergence as a function of the number of samples $N$. With a small number of samples, the NES algorithm can become stuck on local peaks, which leads to a lower probability of success. By increasing the number of exploring samples, we increase the probability of finding the maxima. However, by adding more samples the algorithm requires more computations or consequently more time for the extra samples to converge to the peak. Thus, the decrease in CE and consequent increase in the convergence time can be justified by the guaranteed convergence for some higher number of samples.

## APPLICATION: INTELLIGENT ROBOTIC ARM

Knuth et al. [4, 6] have previously demonstrated autonomous Bayesian exploration using a robotic arm to locate a bright circle in the dark field. There are three parameters: two for the center coordinates and one for the radius. The recorded data consists of light intensity measured at a particular point in the field. The intensity reading is high if the sensor is inside the circle; and low if the sensor is outside the circle. Once a measurement is made, the robot can update its inferences about the center coordinates and the radius of the circle. Such inferences are performed by using the nested sampling algorithm [14] which yields a set of posterior samples, each representing a probable circle.

Given the posterior samples, the design strategy is to optimally select the next measurement location. Every potential measurement point on the field represents a candidate experiment so that experiments are parameterized by the coordinates of the point. Each posterior sample, which represents a hypothesized circle, can be used to predict the intensity measurement for the candidate experiment using an appropriate forward model. This distribution of predicated intensities is then used to compute entropy for every experiment, or equivalently, a location in the field. The optimal experiment is given by the location with the greatest entropy.

Here we illustrate a comparison of the performance of the brute force entropy search employed in [4, 6] to the *NES* algorithm. We take a set of hypothesized circles sampled from the posterior in the inference step and use them to generate a distribution of predictions that give rise to an entropy field. These circles are shown in the two plots in Figure 3. On the left we see the entropy map found using the brute force method, which samples the entire entropy space. This can be compared to the same map sparsely sampled by the *NES* algorithm on the right. The measurement locations that maximize the entropy are indicated by the white regions on the left and the outlined cells on the right.

The brute force technique requires that entropy be computed at every point in the field, so in this problem the number of computations increases with the square of the linear dimensions and resolution of the problem. Use of *NES* avoids the need to compute the entropy for all candidate points. For 25 samples the compression efficiency is found to be 4.35 so that NES only searches at 855 points out of 3721 total candidate points saving about 77% of computations required by brute force method.

## RESULTS AND DISCUSSIONS

For model based learning, every search location corresponding to the parameter space corresponds to a unique candidate experiment. Finding the optimal experiment requires that we compute the utility or cost of conducting such experiments. In this paper we discussed an entropy-based search algorithm for selecting an optimal experiment. We tested the convergence of the algorithm in two-dimensional landscapes defined by the mixture of Gaussians and also studied the probability of success for given landscape with varying number of exploring samples. Increasing more samples increases the probability of success; while at the same time, it also increases the number of computations required to find the global solutions. By computing the rate of success for various numbers of samples, we can get some idea about the optimal number of samples required for convergence.

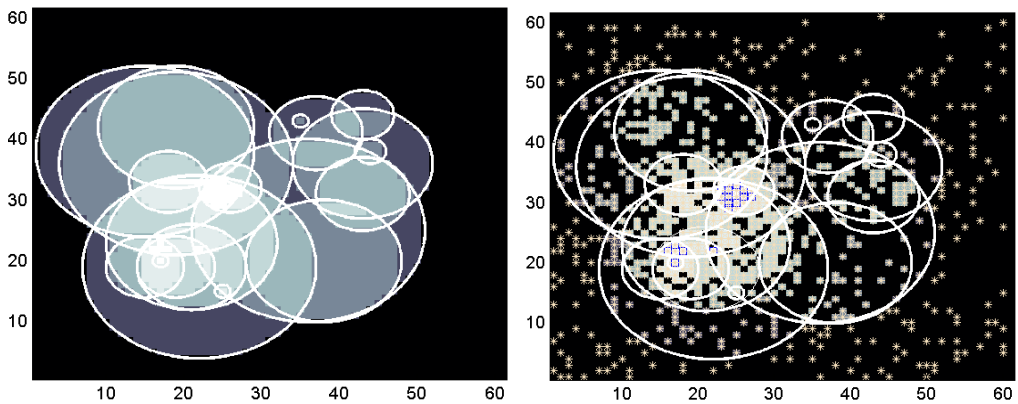

**FIGURE 3** (Left). The sampled circles are illustrated and regions of the two-dimensional space are shaded according to the resulting entropy . Darker color corresponds to the less entropy values. (Right) Demonstration of an NES run. The outlined squares indicate coordinates where the entropy is maximal. Note that samples have converged to more than one candidate experiments.

We also implemented the NES algorithm for the simulation of robotic arm developed at Knuthlab [4, 6] where the algorithm works within the query phase of the experimental design. During the simulations it was found to reduce the number of required computations by about seventy seven percentage. Future work will consider comparing NES algorithm with other available optimization algorithms.

## ACKNOWLEDGMENTS

The authors would like to thank Ariel Caticha and Tom Loredo for informative discussions and Phil Erner for the comments on the text. This work is funded in part by a University at Albany Faculty Development Award.